\documentclass[a4paper,conference]{IEEEtran}

\usepackage{booktabs}
\usepackage{graphicx}
\usepackage{xcolor}
\usepackage{multirow}
\usepackage{enumerate}
\usepackage{amsthm}

\definecolor{red}{HTML}{e74c3c}
\definecolor{green}{HTML}{2ecc71}

\usepackage{amsmath,bm}
\usepackage{graphicx}
\usepackage{dcolumn}
\usepackage[english]{babel}
\usepackage{subcaption,soul}
\usepackage{caption}
\usepackage[numbers,sort&compress,sectionbib]{natbib}
\usepackage[pagebackref=false,breaklinks=true,letterpaper=true,colorlinks=true,linkcolor = black, urlcolor = black,citecolor=black,bookmarks=true]{hyperref}
\usepackage{cleveref}
\usepackage{url}
\usepackage{multirow}
\usepackage{booktabs,comment,balance}
\usepackage{amsfonts}

\newcommand{\calL}{\mathcal{L}}

\ifCLASSINFOpdf
\else
\fi
\begin{document}
%


\title{Arbitrary Shape Text Detection using Transformers}

\author{\IEEEauthorblockN{Zobeir Raisi, Georges Younes, \textrm{and} John Zelek}
\IEEEauthorblockA{University of Waterloo, Waterloo, ON, Canada, N2L 3G1\\
\{zraisi, gyounes,  jzelek\}@uwaterloo.ca}
}


%


\maketitle

\begin{abstract}

Recent text detection frameworks require several handcrafted components such as anchor generation, non-maximum suppression (NMS), or multiple processing stages (\textit{e.g.} label generation) to detect arbitrarily shaped text images.
In contrast, we propose an end-to-end trainable architecture based on Detection using Transformers (DETR), that outperforms previous state-of-the-art methods in arbitrary-shaped text detection.

At its core, our proposed method leverages a bounding box loss function that accurately measures the arbitrary detected text regions' changes in scale and aspect ratio. This is possible due to a hybrid shape representation made from Bezier curves, that are further split into piece-wise polygons. The proposed loss function is then a combination of a generalized-split-intersection-over-union loss defined over the piece-wise polygons, and regularized by a Smooth-$\ln$ regression over the Bezier curve's control points.

We evaluate our proposed model using Total-Text and CTW-1500 datasets for curved text, and MSRA-TD500 and ICDAR15 datasets for multi-oriented text, and show that the proposed method outperforms the previous state-of-the-art methods in arbitrary-shape text detection tasks.


\end{abstract}


%
\IEEEpeerreviewmaketitle

\section{Introduction}
\label{sec:intro}
Scene text detection is the process of accurately localizing text instances in wild images; it is an essential component that enables various practical applications such as text recognition, blind navigation, and topological mapping to name a few \cite{survey_text_21,zobeir_2020}.
While recent text detection methods \cite{deng2018pixellink,liao2017textboxes,liao2018textboxes++,zhou2017east,deng2019_stela,RYOLO_2021,DCLnet_2021} have shown reliable performance on horizontal and multi-oriented text, accurate detection of texts in  an arbitrary geometric layout is still an open-ended problem.

The majority of State-Of-The-Art (SOTA) arbitrary shape text detectors are built on object detection or segmentation frameworks, and can be categorically divided into two classes: segmentation-based \cite{deng2018pixellink,long2018textsnake,wang2019_PAN,PSENet_wang2019,LOMO_2019,textfield_2019,baek2019craft} and regression-based \cite{liao2018textboxes++,zhou2017east,LOMO_2019,textray_2020,abcnet_2020,Furier_2021,abcnetV2_2021}.
The segmentation-based methods \cite{deng2018pixellink,long2018textsnake,wang2019_PAN,PSENet_wang2019,textfield_2019,baek2019craft,DB_2020,DCLnet_2021} encode text instances at a pixel level, and aggregate the resulting pixels to generate a segmentation mask per text instance.
While they are flexible in detecting arbitrarily shaped texts, they require complex architectures and computationally expensive post-processing steps to be able to detect quadrilateral and curved text instances. This results in a high inference time, and increased difficulty to train them, which in turn requires extensive amounts of training data.

On the other hand, regression-based methods \cite{liao2018textboxes++,zhou2017east,LOMO_2019,textray_2020,abcnet_2020,Furier_2021,abcnetV2_2021} are inspired from generic object detection frameworks \cite{fasterrcnn_ren2015,Unet_2015,liu2016ssd,Maskrcnn_He2017,Lin_2017_FPN}, and model text instances as objects. Unlike segmentation-based methods, they output bounding boxes around the text regions using relatively simple architectures; as such, they are fast and easy to train.
%
While some of these methods can achieve good performance on irregular texts, appropriately formulating anchors to fit arbitrarily-shaped text instances is not a solved problem, and requires post-processing steps (\textit{e.g.}, NMS) to achieve a reliable final detection.

Recent advancements in object detection enabled Transformer frameworks \cite{survey_visualTransforumer,transformers_2020,convolutional_2021} like DETR (Detection Transformer) \cite{detr_2020}
to eliminate the need for many of the existing handcrafted post-processing steps such as anchor generation, and non-maximum suppression (NMS) from the object detection pipeline \cite{fasterrcnn_ren2015,YOLO_2016,liu2016ssd,Maskrcnn_He2017}, all while achieving superior performance.
For example, Raisi \textit{et al.} \cite{Raisi_2021_CVPR}, leveraged the DETR \cite{detr_2020} architecture for multi-oriented scene text detection and achieved SOTA performance in some benchmark datasets.
Nevertheless, DETR has difficulties detecting small objects and suffers from a slow convergence rate. To address these issues, \cite{deformableDETR_2020} introduced a deformable attention module to focus on a sparse small set of prominent key elements, thereby performing better in terms of average precision, and obtaining faster convergence during training. However, \cite{detr_2020,deformableDETR_2020} frameworks can only generate rectangular bounding boxes around the detected objects, and cannot handle arbitrarily shaped texts.

In contrast to \cite{detr_2020,deformableDETR_2020}, we propose an end-to-end Transformer-based object detection architecture that can directly localize multi-oriented or curved text instances in the given image. 
Our proposed text representation is tailored to the scene text detection task as it predicts 8 or 16 control points of a quadrangle box or Bezier curve respectively, for each text region; this allows our method to overcome the drawbacks of directly deploying a generic object detector as in \cite{detr_2020} {that predicts only 4 points of every rectangular box.} 

Our main contributions can be summarized as follows:
(1) We propose an end-to-end trainable Transformer-based framework for arbitrary shaped text detection; the proposed architecture can directly output fixed vertices for the Bezier curves that bound multi-oriented and curved text shapes. 
{This is achieved by modifying the prediction head of the baseline pipeline via  designing a new text detection technique that aims to infer $n$-vertices of a polygon or the degree of a Bezier curve that is better suited for irregular-text regions}; and
(2) We propose a loss function that is accurate in measuring the changes in scales and aspect ratios of the detected text regions, and accepts arbitrary shapes of text instances using both Bezier curves and polygon bounding boxes.
(3) {We study the effect of  different vertices of polygon representation with the Transformer's architecture on arbitrary shape text instances.}

\section{Related Work}
\label{sec:related_work}
\subsection{Segmentation-based Methods}
Segmentation-based methods typically decompose text instances in a given image into pixels/segments that are then aggregated into an output mask. Segmentation methods cover a large body of research including \cite{deng2018pixellink,long2018textsnake,wang2019_PAN,PSENet_wang2019,textfield_2019,baek2019craft,DB_2020,DCLnet_2021} to name a few.
For example, PixelLink \cite{deng2018pixellink}, adopted a segmentation framework of SSD \cite{liu2016ssd} with a FCN \cite{long2015fully} to predict the relationship links between pixels of text and non-text instances, to localize similar adjacent pixels, and to group them.
TextSnake \cite{long2018textsnake} proposed to detect the arbitrary shape of text instances with ordered disks and text centre lines.
To efficiently separate close text instances,
PAN \cite{wang2019_PAN} made use of an efficient instance semantic segmentation framework that selectively aggregates text pixels according to their embedding distances, resulting in a model that can handle arbitrary shape text regions.
PSENet \cite{PSENet_wang2019} expanded the final local segmented areas from small kernels to predefined scales, allowing close text instances to be separated using a progressive scale algorithm.
%
TextField \cite{textfield_2019} deployed a deep direction field approach to generate candidate text parts, and to link neighboring pixels.
Different from mentioned word-level detectors, CRAFT \cite{baek2019craft} proposed to detect and connect character regions to generate polygons of arbitrary-shape text instances; this was achieved by training a U-Net \cite{Unet_2015} type framework in a semi-weekly supervised learning process.

\subsection{Regression-based Methods}
Regression based methods such as \cite{liao2018textboxes++,zhou2017east,LOMO_2019,seglinkplus_2019,ddrg_2020,textray_2020,abcnet_2020,Furier_2021,abcnetV2_2021} are mostly inspired by general object detectors (\textit{e.g.}, Faster R-CNN \cite{fasterrcnn_ren2015} and SSD \cite{liu2016ssd}); they directly regress the entire word or text-line with arbitrary shape in an image at object level.

Early regression-based methods such as TextBoxes++ \cite{liao2018textboxes++} and EAST \cite{zhou2017east} used SSD's \cite{liu2016ssd} architecture to detect text regions with rotated rectangles or quadrilateral descriptions.
More recently, \cite{Raisi_2021_CVPR} extended DTER's \cite{detr_2020} architecture to output rotated rectangular boxes directly and achieved SOTA performance in multi-oriented benchmark datasets.
However, these representations ignore the geometric traits of the arbitrary shape of curved texts and end up producing considerable background noise.

To better fit arbitrary shaped text, more advanced methods proposed the use of polygons;
For example, LOMO \cite{LOMO_2019} took advantage of both segmentation and regression-based architectures by utilizing Mask-RCNN \cite{Maskrcnn_He2017} as their base framework, and introducing iterative refinement and shape expression modules to refine bounding box proposals of irregular text regions.
TextRay \cite{textray_2020}, leveraged the SSD framework by eliminating the anchor design, and detecting polygons in the polar coordinate system to better represent arbitrary shape text instances.
ABC-Net \cite{abcnet_2020,abcnetV2_2021} build on a ResNet-50 \cite{ResNet_He2015L} feature extractor with a Feature Pyramid Network (FPN) \cite{Lin_2017_FPN} as their backbone, and introduce a Bezier curve representation in order to detect multi-oriented and curved scene text instances.
FCENet \cite{Furier_2021} extends the base network of \cite{abcnet_2020} by performing some post-processing steps like Inverse Fourier Transforms (IFT) and NMS to reconstruct text contours of arbitrary-shape text instances.


\section{Methodology}
\label{sec:method}
Our proposed framework leverage the efficient and fast-converging encoder-decoder as the base detection architecture \cite{deformableDETR_2020}. A CNN backbone extracts first multi-scale feature maps from the input. After attaching positional encodings to the resulted features, they fed into the Transformer encoder, which outputs refined multi-scale features. 
Then A fixed small set of learnable embedding called object queries is passed through the Transformer decoder parallelly. 
The decoder generates instance-aware query embeddings, which are then fed into a prediction head that directly converts the decoders' outputs into each query's class and bounding box set.
The proposed network is trained by a Bipartite matching loss that utilizes the Hungarian matching algorithm \cite{hungarian_1995} to compare a one-to-one mapping between $N$ queries and $N$ ground-truths \cite{detr_2020}.

In this work, instead of computing $4$ scalars that correspond to the $(x,y,w,h)$ coordinates of the centers ($x,y$) and the height ($h$) and width ($w$) of the box, we extend the number of predicted variables to $2 \times n$ scalars that correspond to the coordinates of the $n$ control points of a Bezier curve in \eqref{eq:Yn} and the $k$ polygon points in \eqref{eq:poly_n_points}.
To train the network, we modify the regression head, along with the loss and matching functions as described in Section \ref{sec:contrib}.



\subsection{Text Regions Representations}
\label{sec:RepresentationTextRegions}

\begin{figure*}[]
    \centering
    \includegraphics[width=\linewidth]{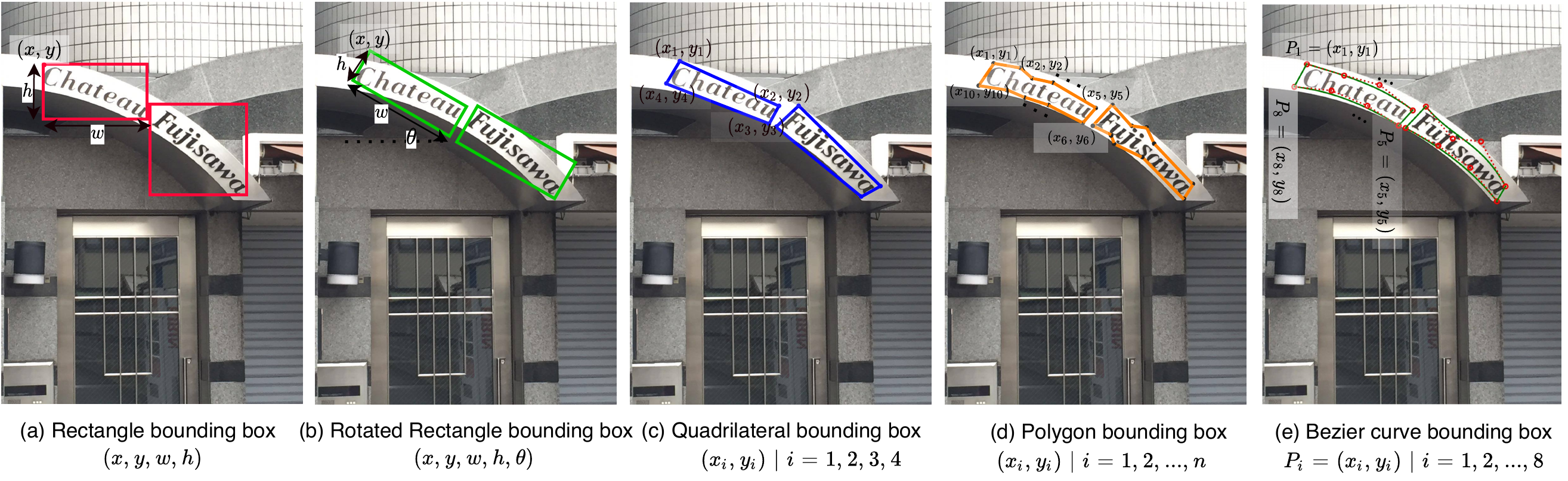}
    \caption{Illustrations of different techniques for representing bounding boxes for scene text detection. The Bezier curves in (e) better draw smooth lines between arbitrary shaped text instances with fixed 8 control points that are more suitable for training our proposed framework. Furthermore, we can better rectify the detected regions in (e), which later lead to a more accurate word recognition performance \cite{abcnet_2020}.}
    \label{fig:typestextregions}
    \vspace{-10pt}
\end{figure*}

\vspace{-3pt}{\flushleft{\textbf{Rectangular Bounding Boxes:}}}
Rectangular bounding boxes are one of the most intuitive representations of horizontal text regions; as shown in Figure \ref{fig:typestextregions}(a), a bounding box $b=[x,y,w,h]^\top$ can encase the text region by simplify defining $(x,y)$ as the bounding box's center point coordinates, and $w, h$ representing the box's width and height respectively.
However, rectangular bounding boxes suffer from several limitations that render them inadequate for irregular text representations;
%
some of these limitations include:
(a) limited ability to distinguish among overlapped or nearby text regions,
(b) they can not precisely bound marginal-text, and
(c) they include large irrelevant background areas that can affect the detector's loss function during training.
%
%
To address these limitations, arbitrary shaped text regions are typically represented using other categories of bounding boxes as shown in Figure \ref{fig:typestextregions}(b)-(d). 

\vspace{-3pt}{\flushleft{\textbf{Quadrilateral Representation:}}}
A Quadrilateral bounding box can be described as $b =[x_1,y_1,x_2,y_2,x_3,y_3,x_4,y_4] ^\top$, where $(x_i,y_i)$ are the four vertices of the quadrilateral arranged in a clockwise order.
The added dimensions allow the quadrilateral to precisely represent various types of text regions including horizontal, multi-oriented, and slight-round texts.

\vspace{-3pt}{\flushleft{\textbf{Polygon Representation:}}}
Polygons are a natural extension of quadrilaterals, where the number of points is increased from 4 to $n-point$ vertices; the bounding box defined by the polygon vertices can then be defined as $b =[(x_i,y_i)|i=1,2,...,n] ^\top$, which can essentially better follow the boundary of a text region, and accordingly represent any arbitrarily-shaped text.

\vspace{-3pt}{\flushleft{\textbf{Bezier Curves:}}}
\label{sec:bezierCurve}
Unlike polygons, a Bezier curve is a parametric curve of degree $n$, $Y_n(t)$, which is used to draw smooth lines between text bounds.
The general form of an $n$-degree Bezier curve can be expressed in terms of a set of $n+1$ control points $\{P_i\}_{i=0}^{n}$ as:
\begin{equation}\label{eq:Yn}
    Y_n(t)=\sum_{i=0}^{n} B_{i,n}(t)P_i, \quad \quad 0 \leq t \leq 1
\end{equation}
where $P_i =(x_i,y_i | i=0,1, \ldots, n)$, $t$ is a normalized independent variable that is used to move along the Bezier curve with a step that determines the smoothness of the curve, and $B_{i,n}(t)$ denotes the $i$th version of the $n-$degree Bernstein Polynomials \cite{bernstein_2013} that are defined using:
\begin{equation}\label{eq:BP}
    B_{i,n}(t)= {n \choose i} t^i (1-t)^{n-i}, \quad \quad i=0,1, \ldots, n
    \vspace{-5pt}
\end{equation}
and ${n \choose i}$ is the Binomial coefficient.

While a $3^{\text{rd}}$-degree Bezier curve, defined by $4$ control points, is effective in representing one side of an arbitrary shape text, another $3^{\text{rd}}$-degree Bezier curve is needed the represent the opposite side (as shown in Figure \ref{fig:typestextregions}(e)), bringing the total number of control points needed to fully represent text boundaries to $8$. The $8$ control points are then computed during regression and prediction as:
\begin{equation}
\label{eq:bezier_points}
    (P_{ij}=x_{ij},y_{ij} | i=0,1,...,3, j=0,1)
\end{equation}
where $b_i$ in \eqref{eq:bloss} are the vertices of the Bezier curve obtained using \eqref{eq:Yn}.

\subsection{Proposed System}
\label{sec:contrib}

\begin{figure}
    \centering
    \includegraphics[width=\linewidth]{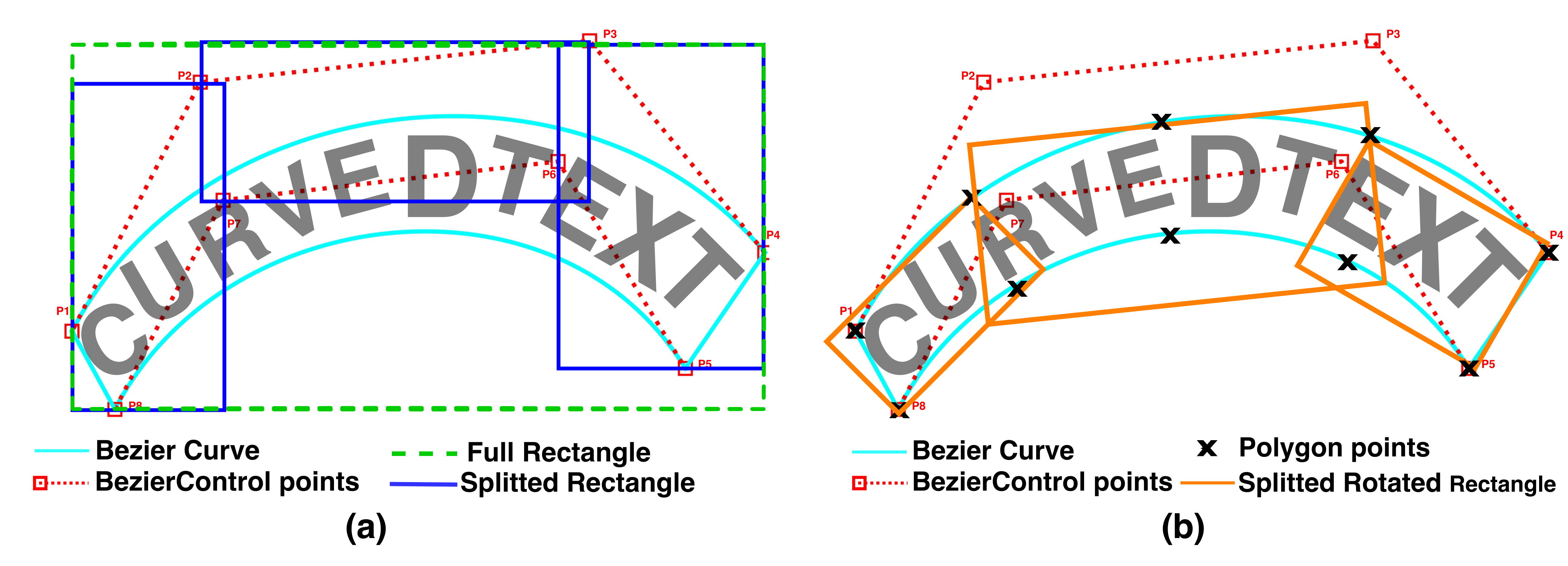}
    \caption{Illustration of the proposed methods. The control points (dotted lines) in (a) and polygon vertices ('x' points) in (b) are predicted directly by the network. The entire rectangle (green dash lines) in (a) is used for Full GIoU calculation. The three split rectangles (blue lines in (a)) and rotated rectangles (orange lines in (b)) make the GIoU and then the Bezier curves (cyan line) and polygon vertices to better bound to high curved text instances.}
    \label{fig:splitted}
    \vspace{-5pt}
\end{figure}

Similar to \cite{abcnet_2020}, we adopt Bezier curves to represent the boundaries of arbitrary shape text instances. To achieve this, we modify the prediction head of deformable DETR's  architecture \cite{deformableDETR_2020} to output 16 parameters that represent the Bezier control points.
However, unlike \cite{detr_2020} and \cite{deformableDETR_2020} that use a generic Generalized Intersection over Union (GIoU) with $\ell_1$-regression \cite{Giou_2018_CVPR} (shown in Figure \ref{fig:typestextregions}(a)),
we propose a split GIoU loss for Bezier control points of \eqref{eq:bezier_points} (shown in Figure \ref{fig:splitted}), along with a Smooth-$\ln$ regression based loss \cite{Raisi_2021_CVPR}. 

The intuition behind the split GIoU is to better compute the difference (loss) between the ground truth and estimated text boundaries. While GIoU can be computed over the Bezier curves, it is computationally inefficient and more complex to calculate the area of intersection between two Bezier curves.
To mitigate this, we split the Bezier curve computed from the regressed control points into several rectangles. The piece-wise GIoU over the rectangles can then be computed efficiently, and the overall set of rectangles defining one text instance are smoothed with the regression loss function over the Bezier curve control points.

The bounding box loss function of \cite{detr_2020} uses a linear combination of $\ell_1$ and GIoU loss.
Let $\hat{b}_i$ and ${b_j}$ denote the $i^{th}$ predicted and $j^{th}$ ground truth bounding boxes, respectively, then we define our loss function as:
 \begin{equation}
     \mathcal{L}_{\textrm{box}}^B({\hat{b}_i, b_j}) =  \lambda_{1} \mathcal{L}_{reg}^B(\hat{b}_i,b_j)+ \lambda_{\textrm{2}}\mathcal{L}_{\textrm{GIoU}}^B({\hat{b}_i, b_j})
     \label{eq:bloss}
     \vspace{-5pt}
 \end{equation}
where $\lambda_{\textrm{1}}$ and $\lambda_{2}\in \mathbb{R}$ are hyper-parameters, and  $\mathcal{L}_{reg}^B(\cdot)$ and $\mathcal{L}_{\textrm{GIoU}}^B(\cdot)$ are the Bezier-curved loss functions based on regression and GIoU.
For regression, we use the Smooth-$\ln$ based Regression Loss as in \cite{Raisi_2021_CVPR}. 
The regression loss is then defined as:
\begin{equation}\label{eq:L_reg}
    \mathcal{L}_{reg}^B(\hat{b}_i,b_j) = (|\Delta b_{ij}| + 1)\ln (|\Delta b_{ij}| + 1) - |\Delta b_{ij}|
\end{equation}
where $\Delta b_{ij} = \hat{b}_i-b_j$ and $|\cdot|$ demonstrates the absolute operator.
The second part of \eqref{eq:bloss} consists of GIoU loss, which plays an important role in the framework of detection using Transformers \cite{detr_2020}.
The GIoU loss
is computed as:
\begin{equation}\label{eq:L_giou}
\calL_{\text{giou}}^B({\hat{b}_i, b_j}) = 1- \text{GIoU}({\hat{b}_i, b_j}),
\end{equation}
The GIoU for two arbitrarily bounding boxes  $\hat{b}_i, b_j\subseteq\mathbb{S}\in\mathbb{R}^n$ can be defined as follows:
\begin{align}
    \text{GIoU}({\hat{b}_i, b_j}) =& \text{IoU}({\hat{b}_i, b_j}) - \scalebox{1.06}{\text{$\frac{\mathbf{Area}(C\backslash(\hat{b}_i\cup b_j))}{\mathbf{Area}(C)}$}} \label{eq:giou}\\
\text{with \quad}
\text{IoU}({\hat{b}_i, b_j}) =& \frac{\mathbf{Area}(\hat{b}_i\cap b_j)}{\mathbf{Area}(\hat{b}_i\cup b_j)}, \label{eq:IoU}
\end{align}
where $C$ shows the smallest area that encloses both prediction and ground-truth boxes $\hat{b}_i$ and $b_j$, and $\mathbf{Area}(\cdot)$ denotes the area of a set.
To compute the GIoU loss for 16  Bezier points of the architecture, we start by calculating the rectangular bounding box that bound all control points of \eqref{eq:bezier_points} in the ground truth and prediction outputs of the network. To better fit to high curved text instances in arbitrary shape benchmarks \cite{ch2017total,CTW_1500_yuliang2017}, we then split the Bezier control points into
several axis-aligned rectangular bounding boxes, where the first rectangular box is computed from $P_1, P_2, P_7, P_8$ Bezier control points, the second and third boxes are also obtained from $P_2, P_3, P_6, P_7$ and $P_3, P_4, P_5, P_6$, respectively.
This process is summarized in Figure \ref{fig:splitted}(a).


\subsection{$n-$point Polygon Ground Truth Generation}
The Bezier control points move outside of the image when the text appears near the margin of an image, requiring negative values of $(x,y)$. Since the final prediction head of \cite{detr_2020,deformableDETR_2020} only outputs positive values, it fails to precisely detect the mentioned text instances. To address this issue, instead of using the Bezier control points directly as shown in Figure \ref{fig:splitted}(a), we first calculate the $3^{\text{rd}}$-degree Bezier curve for each side of the text, defined by $4$ control points. We then recalculate the $n-$polygon vertices (as illustrated in Figure \ref{fig:splitted}(b)) by uniformly sample $n_v$ points as follows:
\begin{equation} \label{eq:poly_n_points}
    p_k=\sum_{i=0}^{n=4} P_iB_{i,n} k / n_v,
\end{equation}
where $p_k$ demonstrates the new $k$-th sampled polygon points, $P_i$ indicates the $i$-th Bezier control points and $n_v$ shows the polygon points used for sampling.
$B_{i,n}$ represents the $n-$degree Bernstein Polynomials \cite{bernstein_2013} as described in \eqref{eq:BP}.
\section{Experimental Evaluation}
\label{sec:experimental_result}
We evaluate the performance of our proposed system, on public scene text detection datasets \cite{karatzas2015icdar,icdar2017,ch2017total,CTW_1500_yuliang2017} that cover a wide range of challenging scenarios. We also perform a set of quantitative and qualitative experiments to benchmark the SOTA text detection \cite{shi2017detecting,zhou2017east,deng2018pixellink,long2018textsnake,textdragon_2019,textfield_2019,PSENet_wang2019,seglinkplus_2019,LOMO_2019,baek2019craft,wang2019_PAN,ddrg_2020,textray_2020,abcnet_2020,Furier_2021,Contournet_2020,DB_2020,abcnetV2_2021,DCLnet_2021} techniques against our proposed model.
Following the criteria used in  \cite{abcnet_2020} to evaluate performance on arbitrary shaped text, and the evaluation metrics  \cite{karatzas2015icdar,icdar2017} used to evaluate ICDAR's multi-oriented text, we report on the Recall, Precision, and H-mean of the various methods.



\subsection{Implementation Details}
\label{sec:implement}
%
We adopt the recent Deformable DETR \cite{deformableDETR_2020} model with a ResNet-50 \cite{ResNet_He2015L} backbone as our base object detector architecture. The number of object queries are set to 300 and an AdamW \cite{adamw_2017} optimizer is used to optimize the parameters of the model. We use a horizontal flip and  {and resize the images similar to} \cite{deformableDETR_2020} for augmentation.
%
All our proposed models are pre-trained on synthetic datasets as in \cite{abcnet_2020} for $20$ epochs with a batch size of 2 per GPU using 4 Tesla V100 GPUs with a learning rate (LR) of $1 \times 10^{-4}$. We follow \cite{deformableDETR_2020} for other hyper-parameters during pre-training.
%
During fine-tuning, we adopt a different LR schedule and train for about $200$ epochs for both the Total-text and CTW-1500 datasets, and drop the LR by a factor of $10$ after $70$ epochs. 
As for ICDAR15, we further pre-train the models using about $10,000$ images of ICDAR17 \cite{icdar2017} dataset for $50$ epochs and then fine-tune for about $300$ epochs to ensure the training converges. 
For calculating the rotated version of bounding box loss function, we used the method described in \cite{Raisi_2021_CVPR}.
\subsection{Datasets}
\label{sec:datasets}


We make use of several recently published and challenging datasets, that can be categorized into multi-oriented text datasets, ICDAR15 \cite{karatzas2015icdar} and MSRA-TD500 \cite{MSRATD500_2012} with quadrilateral representation (Figure \ref{fig:typestextregions}(c), and arbitrary-shaped text datasets, Total-Text \cite{ch2017total} and CTW-1500 \cite{CTW_1500_yuliang2017} with $n$-vertices polygon representation as shown in Figure \ref{fig:typestextregions}(d).

\subsection{Comparisons with SOTA Methods}
\label{sec:comparison}
In this section, we first compare the proposed model with the SOTA methods \cite{baek2019craft,zhou2017east,wang2019_PAN,PSENet_wang2019,deng2018pixellink} on two popular datasets containing curved text: Total-Text
\cite{ch2017total} and CTW-1500 \cite{CTW_1500_yuliang2017}.
We evaluate the datasets on two models: (1) that uses $16$ control points of the Bezier curve with three splits rectangularly (Figure \ref{fig:splitted}(a)) and (2) that uses $20$-points polygon with three splits rotated rectangularly (Figure \ref{fig:splitted}(b)).
\vspace{-3pt}{\flushleft{\textbf{Arbitrary-Shape Text Datasets:}}}
We first compare our baseline and proposed models on two popular benchmarks, Total-Text  and CTW-1500,  containing curved text and have $n$-vertices polygon annotations.
\begin{table*}[]
    \centering
    \caption{Comparison of the detection results on Total-Text, CTW-1500, ICDAR15, and MSRA-TD500 datasets with recent regression and segmentation based methods. The best performance is highlighted in \textbf{bold}.}
\resizebox{\linewidth}{!}{%
    \begin{tabular}{l|c|c|c|c|c|c|c|c|c|c|c|c}
    \hline
    \multicolumn{1}{c|}{ \multirow{2}*{\textbf{Methods}} }& \multicolumn{3}{c|}{\textbf{Total-Text}} & \multicolumn{3}{c|}{\textbf{CTW-1500}} & \multicolumn{3}{c}{\textbf{MSRA-TD500}}&\multicolumn{3}{|c}{\textbf{ICDAR15}}\\
        \cline{2-13}
        &\textbf{Recall}&\textbf{Precision}&\textbf{H-mean}&\textbf{Recall}&\textbf{Precision}&\textbf{H-mean}&\textbf{Recall}&\textbf{Precision}&\textbf{H-mean}&\textbf{Recall}&\textbf{Precision}&\textbf{H-mean}\\
        \hline
        SegLink \cite{shi2017detecting} &- &- &-&- &- &- &70.0 &86.0 &77.0&76.8 &73.1 &75.0\\
        Textboxes++ \cite{liao2018textboxes++} &- &- &-&- &- &-&- &- &-&78.5 &87.8 &82.9\\
        EAST \cite{zhou2017east}  & 50.0 & 36.2 & 42.0 & 49.7 & 78.7 & 60.4 &67.4&87.3&76.1 &78.3 &83.3 &80.7\\
        TextSnake \cite{long2018textsnake} &74.5 &82.7 & 78.4 &{77.8} &82.7 & 80.1 &73.9 &83.2 &78.3&84.9 &80.4 &82.6\\
        TextDragon \cite{textdragon_2019}&75.7&85.6&80.3&82.8&84.5&83.6&- &- &-&83.7&\textbf{92.4}&87.8\\
        TextField \cite{textfield_2019}&79.9&81.2&80.6&79.8&83.0&81.4&75.9&87.4&81.3&80.0 &84.3 &82.4\\
        PSENet-1s \cite{PSENet_wang2019}&77.9&84.0&80.9&79.7&84.8& 82.2&- &- &-&84.5 &86.9 & 85.7\\
        Seglink++ \cite{seglinkplus_2019}&80.9&82.1& 81.5&79.8&82.8&81.3&- &- &-&80.3&83.7	&82.0\\
        LOMO \cite{LOMO_2019}&79.3&87.6&83.3&76.5&85.7&80.8&- &- & 83.5& \textbf{91.3}& 87.2\\
        CRAFT \cite{baek2019craft}&79.9&87.6&83.6&81.1&86.0&83.5&78.2&{88.2}&82.9&84.3 &89.8 &{86.9} \\
        PAN  \cite{wang2019_PAN}&81.0 &{89.3}  &85.0 &81.2 &{86.4} &83.7 &{83.8} &84.4 &84.1 & 81.9&84.0& 82.9\\
        {DDRG} \cite{ddrg_2020} & {84.9} &86.5  & {85.7}&83.0 &85.9& {84.5} &82.3 &88.0& {85.1} &84.7& 88.5& 86.5\\
        TextRay \cite{textray_2020}           &77.9&83.5 &80.6 &80.4 &82.8  &81.6  &-&- &-&-&- &-\\
        ABC-Net-v1 \cite{abcnet_2020}         &81.3&87.9 &84.5 &78.5  &84.4&81.4  &-& -& -&-&-&-\\
        FCENet \cite{Furier_2021}             &82.5&89.3 &85.8 &83.4 &87.6  &85.5 & -&-  &- &82.6&90.1&86.2\\
        CounterNet \cite{Contournet_2020}   & 83.9 & 86.9 & 85.4 &  {84.1} & 83.7 & 83.9 &-&-&- &{86.1}&87.6&86.9\\
        DB \cite{DB_2020}                 &82.5&87.1&84.7  &80.2&86.9&83.4  &79.2&\textbf{91.5}&{84.9} &82.7 &88.2&85.4 \\
        ABC-Net-v2 \cite{abcnetV2_2021}       & 84.1 &\bf{90.2} & {87.0} & 83.8 & 85.6 & {84.7} &81.3&{89.4}&85.2   &86.0&90.4&\bf{88.1}\\ \hline
        \textbf{Our model-1}     &{85.7}  &89.4  &{87.5}    &84.0 &{88.3} & {86.1} &{84.5} &87.4 & {85.9} &81.5 &{89.3} &85.2\\
        \textbf{Our model-2}      &\textbf{86.4}  &89.1  &\textbf{87.8}  &\textbf{85.3} &\textbf{89.2} &\textbf{87.2}  &\textbf{85.0} &88.1 &\textbf{86.5} &83.1 &90.2 &86.5\\
    \hline
    \end{tabular}
}
    \label{tab:tb3}
    \vspace{-5pt}
\end{table*}

\vspace{-5pt}{\flushleft{\textbf{Results of Total-Text:}}}
As seen in Table \ref{tab:tb3},
both proposed models achieved the best performance in terms of Recall and Precision compared to other segmentation-based and regression-based methods.
{The second model outperformed the first model, overall by $\sim 0.6$.}
The {effectiveness} of our contributions are evident in the qualitative results of Figure \ref{fig:qual_abl} as it demonstrates how the Bezier curve and 20-point polygons estimated by our proposed methods can better fit more challenging arbitrary-shaped text instances.

\begin{figure}[]
    \centering
    \includegraphics[width=\linewidth]{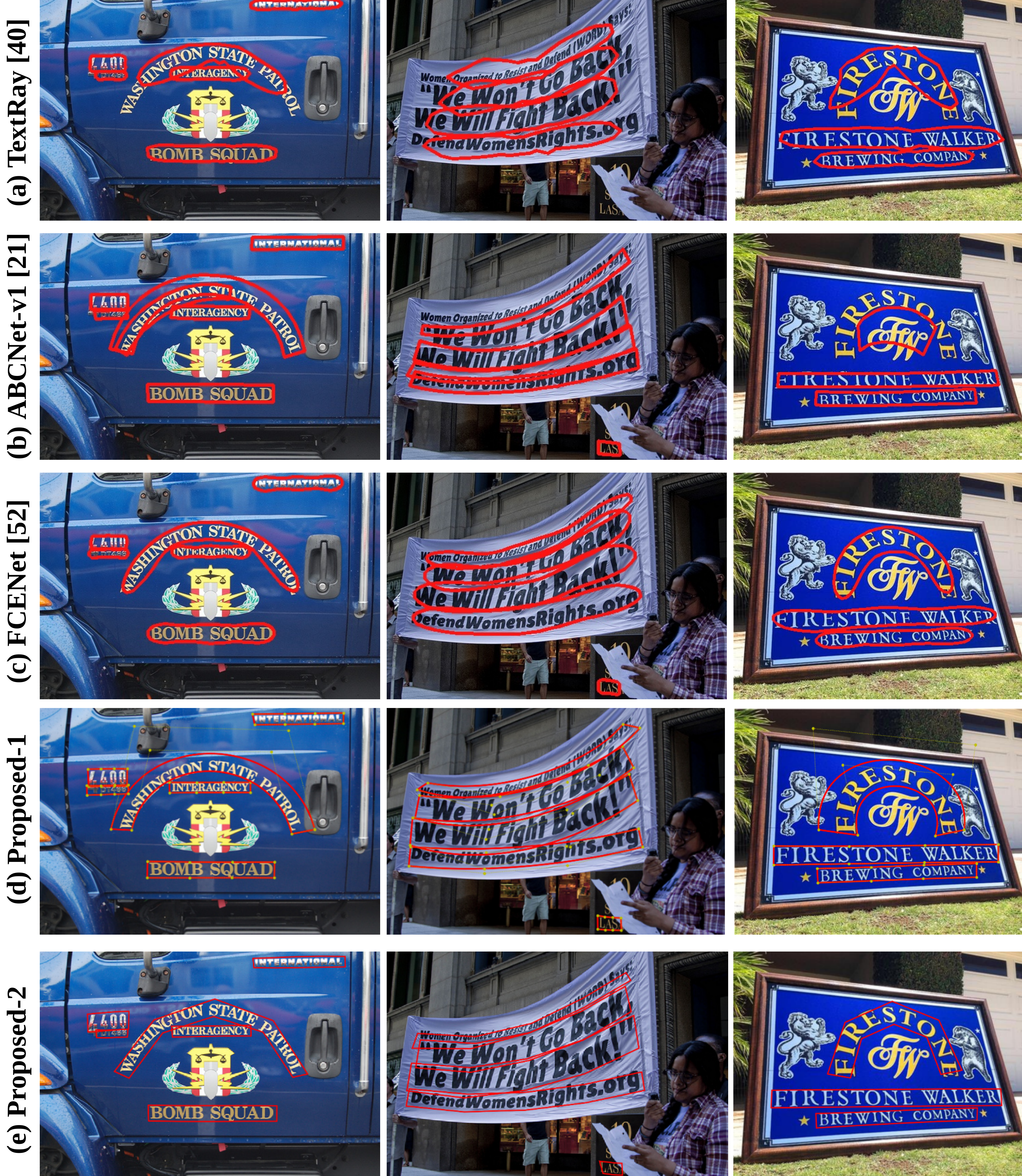}
    \caption{Qualitative comparison of our proposed models among SOTA methods. The sample results of other methods are taken from \cite{Furier_2021}.}
    \label{fig:qual_comp}
    \vspace{-8pt}
\end{figure}

\vspace{-5pt}{\flushleft{\textbf{Results of CTW-1500:}}}
Despite the highly curved text instances in this dataset, our first method surpassed other SOTA systems, achieving the best precision of $88.3\%$ and a H-mean of $86.1\%$. {The second method also performed better than the first on this dataset, which shows how effectively using $20-$points polygon can bound high curved text-line instances.}
The qualitative results using the  proposed methods for some challenging samples of the CTW-1500 \cite{CTW_1500_yuliang2017} dataset are shown in Figure \ref{fig:qual_comp}, where the proposed methods perform better than ABC-Net \cite{abcnet_2020} and TextRay \cite{textray_2020} and exhibit competitive results in some cases against FCENet \cite{Furier_2021} that uses a smoother curve.
The second model that uses 20-points of a polygon with split rotated rectangular outperformed the first model, by overall $\sim 0.6$. It is worth mentioning that the Bezier curve model showed poor performance in detecting text instances near the margin of the images. The second proposed model performed better in these types of text instances.
\vspace{-3pt}{\flushleft{\textbf{Multi-oriented Text Datasets:}}}
We also compare the detection performance of the Transformer's architecture using the Bezier curve for multi-oriented datasets of MSRA-TD500 and ICDAR15. For this purpose,  we use the baseline-$4$ with Smooth-$\ln$ regression and rectangular GIoU loss for training of Bezier curve and 20-points polygon models because of the quadrilateral annotation in these datasets. It is worth mentioning that splitting the GIoU in these datasets does not affect to the final performance. 
\vspace{-3pt}{\flushleft{\textbf{Results of MSRA-TD500:}}}
As shown in Table \ref{tab:tb3} our proposed methods achieves SOTA results in terms of Recall of $85.0\%$ and H-mean of $86.5\%$. The proosed-2 model that uses 20-points polygon representation outperformes the Bezier curve representation and it surpasses the previous best method by a relatively big margin of $\sim 4\%$ and $\sim 1.5\%$ on the Recall and H-mean performances, respectively.
\vspace{-3pt}{\flushleft{\textbf{Results of ICDAR15:}}}
As shown in Table \ref{tab:tb3}, our both models achieve competitive results with SOTA detection models in ICDAR-15 datasets. When using a $20-$points polygon our models outperform the Bezier curve representation with $16$ control points.
\vspace{-5pt}
\subsection{Ablation Study}
\label{sec:ablation}
To assess the added value of the various components in our model, we performed an extensive ablation study on Total-Text and CTW-1500 as demonstrated in 
Table \ref{tab:ablation}.

We started the experiments by eliminating the GIoU loss and training the model with $\ell_1$ loss only; the model achieved a H-mean performance of $79.01\%$ and $78.25\%$ for Total-Text and CTW-1500 datasets, respectively. We then replaced the $\ell_1$ with the Smooth-$\ln$ loss, yielding a slightly improved H-mean.

We found that only using the GIoU loss defined over the entire rectangle led to further performance boosts, which in turn was further improved when we combined both GIoU and Smooth-$\ln$ losses.
Then, we evaluated the split version of GIoU loss with $3$ rectangles achieved the best performance by improving $\sim 4\%$ and $\sim 2.5\%$ for Total-Text and CTW-1500 datasets in the ablation study.

\begin{figure}[t]
    \centering
    \includegraphics[width=\linewidth]{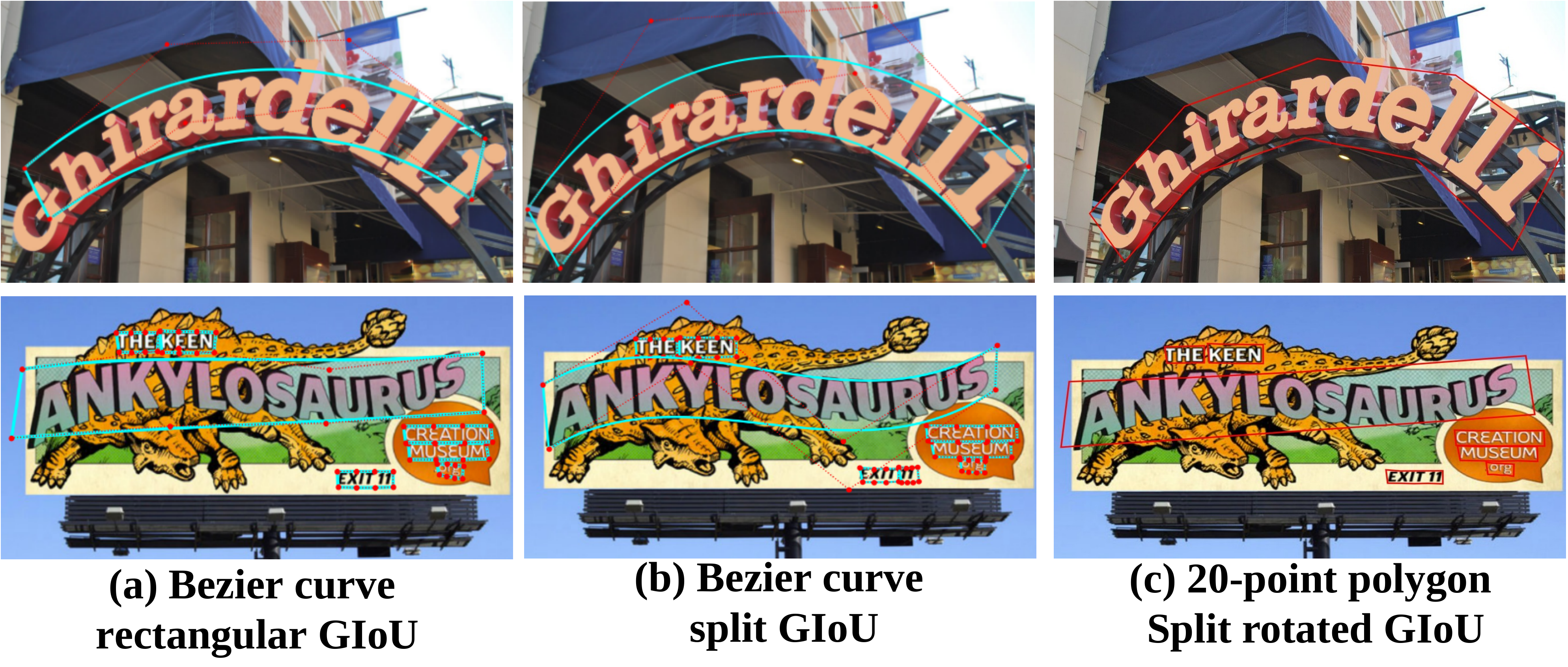}
    \caption{Compare the effect of using split GIoU and baseline GIoU. As seen, the proposed methods with split GIoU  in Table \ref{tab:ablation} better fits the highly curved text instances.}
    \label{fig:qual_abl}
    \vspace{-5pt}
\end{figure}

Finally, we conducted another experiment by using a $20-$points polygon representation with $3$ split rotated rectangles and rotated loss functions as shown in Figure \ref{fig:splitted}(b). Applying this system on the network's head outperformed the first model, especially on the  CTW-1500 dataset by a margin of $\sim 1\%$.
It is worth mentioning that using a split version of the rotated rectangle does not affect the Bezier curves' H-mean performance on the mentioned datasets.
The qualitative results on some challenging cases of Total-Text (shown in Figure \ref{fig:qual_abl}) confirm the effectiveness of the proposed methods with split GIoU when compared to only using a single rectangular GIoU.

\begin{table}[]
    \centering
    \caption{Ablation study on the effects of the various  proposed components on the H-mean metric for Total-Text \cite{ch2017total} and CTW-1500 \cite{CTW_1500_yuliang2017} datasets. R and RR denote the rectangle and rotated-rectangle, respectively.}
    \resizebox{\linewidth}{!}{%
    \begin{tabular}{l|cccccc}
    \hline
        \textbf{Method} &  \textbf{Reg} &\textbf{GIoU} &\textbf{\# split}  & \textbf{Total-Text} & \textbf{CTW-1500}    \\ \hline
        Baseline-1       &\checkmark    &-                  &-             &79.01  &78.25\\
        Baseline-2   &\checkmark    &-                  &-             &79.52  &78.63 \\
        Baseline-3       &   -          &\checkmark         &R(1)             &82.46  &80.83\\
        Baseline-4       &\checkmark    &\checkmark         &R(1)             &83.41  &83.70\\
        \hline
        \textbf{Our model-1}      &\checkmark     &\checkmark   &R(3)  & \textbf{87.50} &\textbf{86.10} \\
        \textbf{{Our model-2}}      &\checkmark     &\checkmark   &RR(3)  & 87.80 & 87.20 \\ \hline
    \end{tabular}
    }
    \label{tab:ablation}
\end{table}

We also trained the Total-Text \cite{ch2017total} dataset with different fixed $8,16,20,24,40,80$-\textit{points} of polygon representation and compared it with Bezier curve representation in Table \ref{tab:poly_bez}. 
The reason for using the Total-text dataset in this experiment is that it contains challenging curved and oriented text instances at the word level.
For a fair comparison, we used a model with similar loss function and split rectangle  in Table \ref{tab:ablation} and the whole training set of Total-text. We trained both models for 300 epochs.
As seen, the Bezier curve with $16$ control points and $20-$points polygon representation are more suitable for detection than using other vertices of a polygon. In addition, we continue experimenting by training the first and second models that use three split GIoU with 16 Bezier control points, and three splits rotated GIoU with 20-point polygon representations, respectively, which the second model performed better in terms of precision and H-mean.
%
\begin{table}[]
    \centering
    \caption{Ablation study of our model using different points of Polygon vs. Bezier (16 points) representation for Totat-Text.}
    \resizebox{0.95\linewidth}{!}{%
    \begin{tabular}{l|cccc}
    \hline
         \textbf{Method} &\textbf{\# points}& \textbf{Recall} & \textbf{Precision} & \textbf{H-mean}\\ \hline
         Bezier curve  & \textbf{16}  & 64.5 & 71.3&67.7 \\
         Polygon & 8 & 51.7&59.6  &55.4\\
         Polygon & 16 &62.0&68.6&65.1\\
         Polygon &\textbf{20} & 64.2&\textbf{73.5}&\textbf{68.5}\\
         Polygon &24&63.6&67.6&65.5\\
         Polygon & 40 &\textbf{64.8}  &59.7  &62.1\\
         Polygon & 80 &20.4  &58.7  &30.3\\
         \hline
         \textbf{Our model-1} & 16&\textbf{66.2} &74.3 &70.0\\
         \textbf{Our model-2} &20&66.1&\textbf{76.6}&\textbf{70.9} \\
         \hline
    \end{tabular}
    }
    \label{tab:poly_bez}
    \vspace{-5pt}
\end{table}

%
\section{Conclusion}
We have presented an arbitrary-shape text detector that directly outputs the bounding boxes of arbitrary shape text instances in natural images. The proposed framework builds on DETR's architecture to output a fixed set of Bezier curve's control vertices and $n-$points of polygon, which in turn can be used to represent arbitrary polygons of curved and multi-oriented texts.
For accurate detection, especially on different challenging arbitrary shape text instances in irregular-text datasets such as Total-Text and CTW-1500, we have also proposed a split version of the Bezier curve and $n-$points of polygon computed from the regressed control points into several rectangles to better fit to the highly curved texts.

We have validated our proposed system using several quantitative and qualitative experiments on challenging benchmark datasets, including multi-oriented quadrilateral annotated text and curved text with $n$-vertex polygons representations. We have also compared the performance of our proposed method with SOTA scene text detection methods, and demonstrated the superior performance of our models on arbitrary shape text and multi-oriented text benchmarks. Our best proposed model that uses a $3$ splits rotated rectangular loss function  achieves the best H-mean performance of $87.8\%$ and $87.2\%$ for Total-Text and CTW-1500 datasets, respectively. Our system also exhibits SOTA performance in Recall ($85.0\%$) and H-mean ($88.1\%$) on the MSRA-TD500 dataset and yield competitive results for ICDAR15 benchmarks.

%


\vspace{-5pt}
\section*{Acknowledgment}
\vspace{-0.2cm}
We would like to thank the Ontario Centres of Excellence (OCE),
the Natural Sciences and Engineering Research Council of Canada
(NSERC), and ATS Automation Tooling Systems Inc., Cambridge,
ON, Canada for supporting this research work.
{
\small
\bibliographystyle{IEEEtran}
\bibliography{uw-ethesis.bib}
}

\end{document}